\def\year{2021}\relax
\documentclass[conference,10pt,letterpaper]{STY/ieeeconf}

\usepackage{cite}
\usepackage{times} % DO NOT CHANGE THIS
\usepackage{helvet} % DO NOT CHANGE THIS
\usepackage{courier} % DO NOT CHANGE THIS
\usepackage[hyphens]{url} % DO NOT CHANGE THIS
\usepackage{graphicx} % DO NOT CHANGE THIS
\usepackage{graphicx} % DO NOT CHANGE THIS
\usepackage{caption}
\usepackage{amsmath,amssymb}

\IEEEoverridecommandlockouts                  
\usepackage[utf8]{inputenc}
\usepackage{amsmath}
\usepackage{amssymb}
\usepackage{bm}
\usepackage{epstopdf}
\usepackage{microtype}
\let\labelindent\relax
\usepackage{enumitem}
\usepackage[normalem]{ulem}
\usepackage[linesnumbered,vlined,ruled]{algorithm2e}
\usepackage{placeins}
\usepackage{makecell}
\usepackage[switch]{lineno}
\usepackage{microtype}

\newcommand{\X}{\mathbb{X}}
\newcommand{\Xcal}{\mathcal{X}}

\newcommand{\Ycal}{\mathcal{Y}}
\newcommand{\Hi}{\mathbb{H}}

\newcommand{\E}{\mathcal{E}}

\DeclareMathOperator{\softmax}{softmax}

\title{A Hierarchical Bayesian Model for Inverse RL in Partially-Controlled Environments}
\author{Kenneth Bogert$^1$ and Prashant Doshi$^2$
\thanks{$^1$Kenneth Bogert is with Department of Computer Science, University of North Carolina, Asheville,
NC 28804 {\tt\small kbogert@unca.edu}}%
\thanks{$^2$Prashant Doshi is with THINC Lab, Dept. of Computer Science, University of Georgia, Athens, GA 30602
{\tt\small pdoshi@uga.edu}}%
}
\begin{document}

\maketitle

\begin{abstract}

Robots learning from observations in the real world using inverse reinforcement learning (IRL) may encounter objects or agents in the environment, other than the expert, that cause nuisance observations during the demonstration. These confounding elements are typically removed in fully-controlled environments such as virtual simulations or lab settings. 
When complete removal is impossible the nuisance observations must be filtered out. However, identifying the source of observations when large amounts of observations are made is difficult.  To address this, we present a hierarchical Bayesian model that incorporates both the expert's and the confounding elements' observations thereby explicitly modeling the diverse observations a robot may receive.  We extend an existing IRL algorithm originally designed to work under partial occlusion of the expert to consider the diverse observations.  In a simulated robotic sorting domain containing both occlusion and confounding elements, we demonstrate the model's effectiveness. In particular, our technique outperforms several other comparative methods, second only to having perfect knowledge of the subject's trajectory.
    
\end{abstract}

%------------------------------------------------------------------------------------------------------
\section{Introduction}
\label{sec:intro}
%------------------------------------------------------------------------------------------------------

Common practice for apprenticeship learning in robotics is to employ a sensor suite to gather information about the observed expert.  This sensor data constitutes a trajectory modeled by a Markov decision process (MDP), which specifies the state of and action taken by the subject agent at successive time steps as it performs the task. Techniques such as classical machine vision and deep learning may be combined to automate this process for complex sensors such as depth cameras~\cite{Soans20:SAnet}. On acquiring the trajectories, they are used by a machine learning method, such as inverse reinforcement learning (IRL)~\cite{Ng00:Algorithms}, to learn a model of the subject’s demonstration so that the apprentice robot may then perform the same task.

While engineers using apprenticeship learning are ultimately concerned with the deployed robot’s performance a significant amount of time and expertise must be spent on developing the portion of the observation system used only during demonstrations. It is this portion which yields the trajectories that are used in the learning. 

One source of difficulty is the inherent noise present in the sensor stream.  Sensors may fail to uniquely indicate the state or action of the subject, may fail temporarily during the demonstration, or the detection is obfuscated due to the presence of confounding elements in the environment.  {\em Confounding elements are objects in the background, other agents moving through the environment, or objects that cause occlusion of the expert.}  Typically these issues are accommodated manually by strictly controlling the environment during demonstrations, repeating demonstrations multiple times, or manually editing noisy sensor streams.  

We present a general hierarchical Bayesian model that captures the uncertainty present in observations during expert agent demonstrations in partially controlled, real-world situations to perform apprenticeship learning. A partially-controlled environment is one in which the set of confounding elements is finite, static, and known.   
Our model exploits the underlying incomplete MDP that is being learned to provide structure to noisy observation data.  We demonstrate our model’s effectiveness in a formative Gridworld scenario and in a larger robotic onion-sorting experiment in which a number of confounding elements are present.

%------------------------------------------------------------------------------------------------------
\section{Preliminaries}
\label{sec:background}
%-----------------------------------------------------------------------------------------------------

IRL connotes both the problem and method by which an agent learns goals and preferences of another agent that explain the latter’s observed behavior~\cite{Russell98:Learning,Ng00:Algorithms}. 
The observed subject agent E is usually considered an expert in the performed task. To model the subject agent, it is assumed that the expert is executing the optimal policy $\pi^E$  of a standard MDP $\langle S,A,T,R \rangle$, with the parameters taking their usual meaning. The learning agent L is assumed to exhibit perfect knowledge of the MDP parameters except the reward function. Therefore, the learner's task is to infer a reward function that best explains the observed behavior of the expert under these assumptions.

The reward function is commonly modeled as a linear combination of $K$ binary features, $\phi_k$:  $S  \times  A$ $\rightarrow$  $[0,1]$ which are known to the learner. 
Then, $R(s,a)  =  \sum_{k=1}^K  \theta_k\cdot \phi_k(s,a)$,  where  $\theta_k$  are  the  {\em  weights}  in  vector $\bm{\theta} $, unknown to the learner.  The learner's task is now reduced to finding this vector. The expert's behavior is provided in the form of a \textit{demonstration}, a  finite, non-empty collection  of trajectories  of varying lengths, $\Xcal{} = \{X^T|X^T \in  \X^T\}$.  $X^T$ is  a trajectory of finite length $T$ assumed to be generated by the MDP attributed  to the expert $\X^T  = \{X|X=(  \langle s,  a \rangle_1,  \langle s,  a \rangle_2,  \ldots, \langle  s, a \rangle_T )\},  \forall s \in S_E,  \forall a \in A_E \}$. Many IRL algorithms make use of \textit{feature expectations} to provide sufficient statistics of a demonstration, defined as $\hat{\phi}_k     =     \frac{1}{|\Xcal|}     \sum_{X     \in     \Xcal{}}
 \sum_{\langle s,a \rangle_t 	\in X} ~$ $\phi_k(\langle s,  a \rangle_t)$.

 To evaluate the quality of a learned reward $R^L$, a popular metric is the {\em inverse learning error} (ILE)~\cite{Choi11:Inverse} obtained as 
 ILE~$= || V^{\pi^E} - V^{\pi^L}||_1$, where $V^{\pi^E}$ is the MDP value function computed using the true reward and the expert's optimal policy for it, $V^{\pi^L}$ is the value function using the true rewards and the optimal policy for the \textbf{learned} reward.

%~~~~~~~~~~~~~~~~~~~~~~~~~~~~~~~~~~~~~~~~~~~~~~~~~~~~~~~~~~~~~~~~~~~~~~~~~~
\subsection{IRL under Occlusion: HiddenDataEM}
\label{subsec:occlusion}
%~~~~~~~~~~~~~~~~~~~~~~~~~~~~~~~~~~~~~~~~~~~~~~~~~~~~~~~~~~~~~~~~~~~~~~~~~~

The problem of IRL is generally ill-posed because for any given behavior there are infinitely-many reward functions which may explain the behavior.  To resolve this, MaxEntIRL~\cite{Ziebart08:Maximum} finds the distribution over all trajectories that exhibits the maximum entropy while matching the observed feature expectations. A generalization,  MaxCausalEntIRL~\cite{Ziebart2010a}, finds stochastic policies with maximum entropy and improves tractability with long trajectories.  However, these algorithms assume complete, error free trajectories are received from the expert which is difficult to provide in many robotic contexts.

Our motivating application involves a collaborative robotic arm sorting onions as they move down a conveyor belt in a processing shed. It learns how to sort by using its sensors to observe an expert worker whose movement may not be fully observed for various reasons such as other persons moving in front of the camera. Previous methods~\cite{Bogert18:Multi} denote this special case of partial observability where some states are fully hidden as \textit{occlusion}. Subsequently, the  trajectories  gathered  by the  learner  exhibit  missing  data associated with  timesteps where  the expert robot  is in one  of the occluded  states.  The  empirical feature expectation of the  expert $\hat{\phi}_k$ will thus exclude the occluded states (and actions in those states).

To ensure feature expectations account for  the missing  data, recent approaches~\cite{Bogert:2017:Scaling,Bogert16:Expectation} take an  expectation over  the missing  data conditioned  on the observations.  
Let $Y$  be the observed  portion of a  trajectory, $H$ is one  way of completing the hidden parts of  this trajectory,  and $X =  Y \cup H$.  Treating $H$ as a latent variable allows us to redefine the expert's feature expectations:
\begin{align}
  \hat{\phi}^{H|_Y}_{\bm{\theta},k}  \triangleq  \frac{1}{|\Ycal{}|}  \sum\limits_{Y
    \in   \Ycal}   \sum\limits_{H   \in   \Hi{}}   Pr(H|Y;\bm{\theta})
\sum\limits_{t=1}^T \phi_k(\langle s, a \rangle_t)
\label{eq:latent-phi}
\end{align}
where $\langle  s,a \rangle_t \in Y  \cup H$, $\Ycal{}$ is  the set of
all  observed   $Y$, $\Hi$ is the set of
all possible hidden $H$ that can complete a trajectory. 
The nonlinear program of MaxEntIRL~\eqref{eq:ziebart-max-ent} is modified to use these new feature expectations.  
\begin{small}
\begin{align}
&  \max \limits_{\Delta} \left( -\sum\nolimits_{X \in \mathbb{X}} Pr(X)~ log~Pr(X) \right ) \nonumber \\
&  \mbox{{\bf subject to}}~~~
  \sum \nolimits_{X \in \mathbb{X}} Pr(X) = 1 \nonumber \\
& E_{\X}[\phi_k]  = \hat{\phi}_{k} ~~~~\forall k
\label{eq:ziebart-max-ent}
\end{align}
\end{small}
\noindent Where $E_{\X}[\phi_k]=\sum \nolimits_{X \in \mathbb{X}} Pr(X)~ \sum_{\langle s,a \rangle_t 	\in X}  \phi_k(\langle s, a \rangle_t)$ and $Pr(X) \propto  e^{ \sum\nolimits_{\langle s,a \rangle_t 	\in X}\bm{\theta}^T  \phi(s,a)}$. Notice that  in the  case of  no occlusion  $\Hi$ is  empty and
$\Xcal{} =  \Ycal{}$.  Therefore $\hat{\phi}^{H|_Y}_{\bm{\theta},k}  = \hat{\phi}_k$.  Thus,  this
method generalizes MaxEntIRL. However, the program becomes nonconvex due to
the presence of $Pr(H|Y)$ and finding its optima by Lagrangian
relaxation  is not  trivial.   Wang et  al.~\cite{Wang12:Latent}
suggests a log  linear approximation %obtain  maximizing $P(X)$ and
that casts  the  problem  as  a likelihood
maximization   that    can   be   solved   within    the   schema   of
expectation-maximization~\cite{Dempster77:EM}.  An application of this
approach to  the problem of  IRL  under occlusion yields a method labeled as {\sf HiddenDataEM}, which consists of the following two steps (with more details in~\cite{Bogert16:Expectation}):

\noindent    {\bf    E-step}    This   step    involves    calculating
Eq.~\ref{eq:latent-phi}  to  arrive  at  $\hat{\phi}^{Z|_Y,(t)}_{\bm{\theta},k}$,  a
conditional  expectation  of  the  $K$  feature  functions  using  the
parameter  $\bm{\theta}^{(t)}$ from  the  previous  iteration. We  may
initialize the parameter vector
randomly.\\
\noindent    {\bf    M-step}    In    this    step,    the    modified
program  is     optimized    by     utilizing $\hat{\phi}^{Z|_Y,(t)}_{\bm{\theta},k}$  from  the  E-step  above  as  the  expert's   feature  expectations   to  obtain   $\bm{\theta}^{(t+1)}$. Our experiments utilized gradient descent~\cite{Steinhardt14:Adaptivity} applied to the program's dual. 

While EM converges but possibly  to local  minima, this  process is  repeated with
random initial $\bm{\theta}$ and the solution with the maximum entropy is chosen as the final one.

%------------------------------------------------------------------------------------------------------
\section{Controlled Environments}
\label{sec:controlled}
%------------------------------------------------------------------------------------------------------

HiddenDataEM assumes that trajectories are perfectly observed with some data missing due to occlusion.  Data acquired from a robot's sensor suite, however, is likely to be noisy and provides partial information about the subject.  We seek, therefore, to generalize the E-step of HiddenDataEM to environmental and sensing noise 
to compute a distribution over possible demonstrations $\Xcal$ given the observations.

We first discuss a specific simplification of our model. It applies to scenarios where the learner (or its operator) has full control over the environment such that there is no possibility of extraneous observations and the learner has perfect knowledge of the state and action of the subject agent at each timestep. This is a popular assumption for IRL, often possible in simulations only, but not practicable in many real-world robotic contexts.

\begin{figure}[!ht]
	\centering
	\includegraphics[width=.2\textwidth]{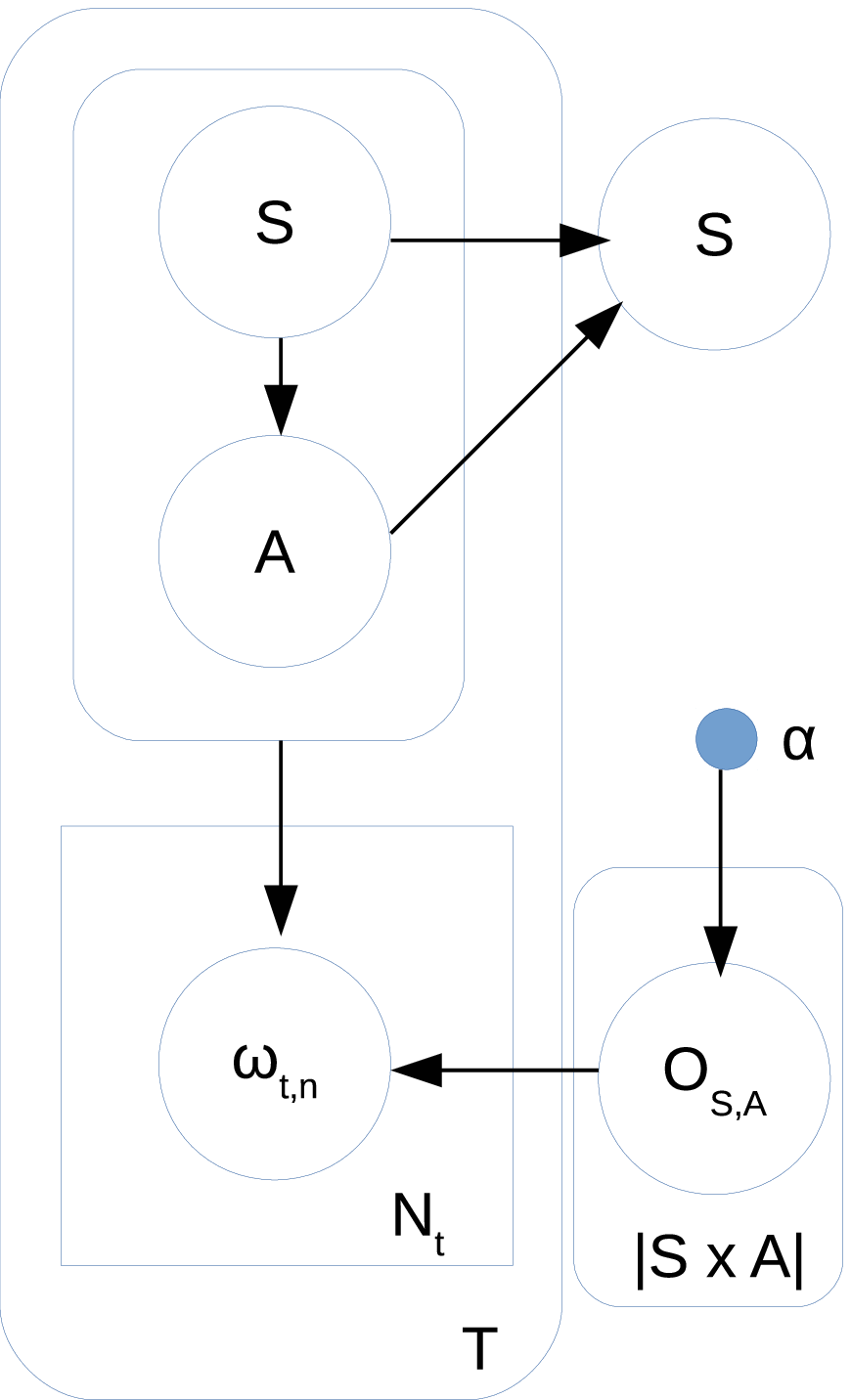}
	\includegraphics[width=.2\textwidth]{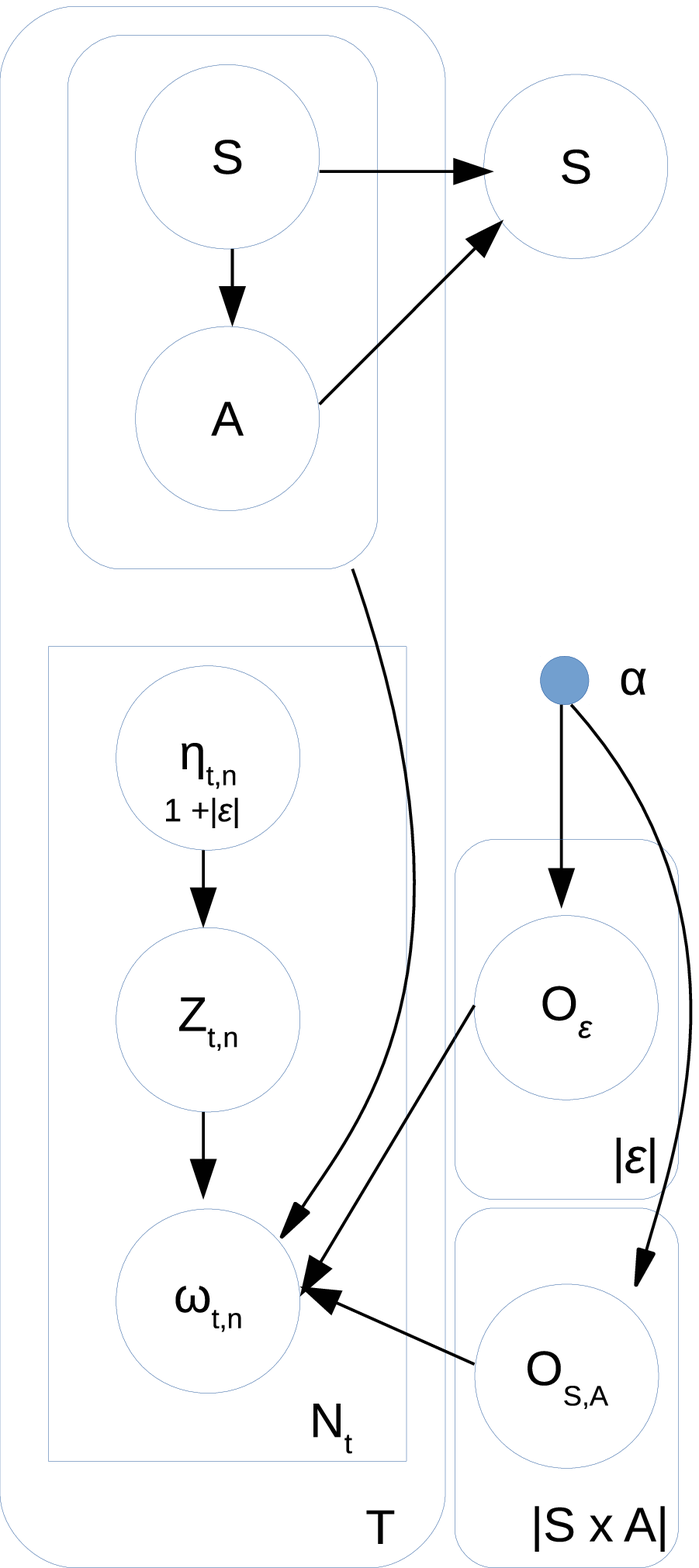}
	\caption{\small {\bf (left)} Controlled environment per-timestep model.  At each timestep $t$, $N_t$ observations $\omega$ are sampled from the observation model $O_{S, A}$. {\bf (right)} Partially controlled environment per-timestep model.  Each observation $\omega_{t, n}$ is labeled with its source, $Z_{t, n}$.  We sample the source from the distribution $\eta_{t,n}$ as it may not be known.}
	\label{fig:controlledtimestepmodel}
	\vspace{-0.1in}
\end{figure}

We emphasize that the sensor data may be markedly different from the state-action trajectory data.  For instance, suppose a video camera is used to record the expert subject and a machine vision algorithm such as SIFT processes the video stream and produces a stream of features, which we call observations. If this camera runs at 30 fps, potentially {\em hundreds to thousands of observations} are generated every second, with many duplicates.  Now, let the MDP timestep correspond to one wall-clock second and the state and action sets are discrete. Then, the many observations are caused by an observation function (sensor model) dependent on the subject’s state and action variable, i.e., $O(s,a, \omega)$, where $\omega \in \Omega$ are sensed observations (features). We show this model graphically in Figure~\ref{fig:controlledtimestepmodel} (left). Of note is that there may be any number of observations at each timestep, $N_t$. This characteristic makes it different from a hidden Markov model in which the state and action of the expert are also hidden but which models a single observation at each timestep.  

It is generally impractical to have complete knowledge of $O$ due to the nature of these complex observations.  Instead, we allow $O$ to be sampled from a Dirichlet distribution with hyperparameters $\alpha$.  Let the likelihood distribution of $\Omega$ be multinomial, which makes the Dirichlet over $O$ a conjugate prior and allows us to set $\alpha$ to be the count of each observation seen per state and action.  

Controlled environments, such as simulations, facilitate this procedure as the state and action of the subject may be perfectly known. This learned model may then be transferred to less controlled environments. However, we may expect the expert to present some significant observation differences in this case.  Therefore, we may scale the  hyperparameters by a small constant to prevent overfitting.

\section{Partially Controlled Environments}

Of course, robotic apprenticeship learning cannot usually be performed in sterile, controlled environments.   Confounding elements often remain in the environment during demonstrations which cause extraneous observations. These observations must be filtered out in order to prevent incorrect trajectories from being produced that will distort  the apprenticeship learning.

This is a problem of identification; if each observation has its source perfectly labeled we could simply exclude all non-subject agent observations and use the controlled environment model from Section~\ref{sec:controlled}.  In practice, however, an observation's source may not be known due to the inherent uncertainty and noise.  One common approach is to restrict the accepted observations in the hope that non-subject agent observations will be removed.  But this increases the chances of occlusion when many of the available observations are not clear. This also precludes using unexpected, opportunistic observations, such as the presence of a mirror or a reflective surface in the environment, as these extra sources may get filtered out.

We take a different approach with the goal of utilizing all available information while identifying and filtering out observations caused by confounding elements.  Suppose we enumerate every confounding element present in the environment during demonstrations (this is reasonable assuming that the environment has been at least partially controlled to limit these elements). We assume that the set of observations received at a given timestep is due to some mixture of these elements and the subject agent.  Let $\E$ be the set of all confounding elements.  To the Bayesian model, we add a label variable $Z$ for each observation $\omega$. $Z$ identifies the source of $\omega$ and may take on a value either from $\E$ or one additional value indicating ``subject agent''.  We expand the observation model to include all elements in $\E$.  Note that when $Z$ labels an $\omega$ as ``subject agent'' the observation model conditional on the subject agent's state and action is chosen, as was the case in our controlled environments model. Thus, $Z$ acts as a multiplexer select input. 

As machine vision is often uncertain about the identification of an observation, $Z$ is itself unknown.  {\em Thus, we model it as being sampled from a distribution $\eta(Z)$, which incorporates any information the observation system has about the source of a given $\omega$ and is of size $1 +|\E|$, with the first entry indicating ``subject agent''.}  This allows for a rich set of information to be incorporated beyond simply that an $\omega$ was observed.  An advanced system could track the movement of agents to help with the identification, it could interpret the observation in the context of others nearby, or express that an observation was vague or unusual.  To illustrate, machine vision based object detection such as using the Python ImageAI library could be employed on a RGB video stream to produce observations. These systems produce a probability distribution of possible object identifications for each detected object, which could be utilized as $\eta$.  We show the new generalized hierarchical Bayesian model in Figure~\ref{fig:controlledtimestepmodel} (right).

\subsection{Finding the Expert's Trajectory Distribution}

Let $\mathbb{O}  = \{  \langle {\bm \omega},  {\bm \eta} \rangle_1,  \langle {\bm \omega},  {\bm \eta} \rangle_2,  \ldots, \langle  {\bm \omega}, {\bm \eta} \rangle_{|\Xcal|} \}$  be the set of all observations produced from the subject's demonstration $\Xcal{}$ each associated with a distribution over their corresponding labels (source distribution).  To compute distributions over trajectories given $\alpha$ and $\mathbb{O}$ we also require the subject's policy $Pr (A | S)$. Unfortunately, as IRL is attempting to learn this distribution, we will not have the correct policy ahead of time.  We resolve this by using the currently found trajectory distributions to compute $\hat{\phi}$ in the E-step of HiddenDataEM, revising Eq.~\ref{eq:latent-phi} to yield Eq.~\ref{eq:ce-phi}.  Using the policy produced by the subsequent M-step we iteratively improve our likelihood estimate of the subject's true demonstration.
\begin{align}
  \hat{\phi}^{\mathbb{O}}_{\bm{\theta},k}  \triangleq   \frac{1}{|\mathbb{O}|} \sum\limits_{\langle {\bm \omega}, {\bm \eta} \rangle \in \mathbb{O}} \sum\limits_{X
    \in   \mathbb{X}}   Pr(X | {\bm \omega}, {\bm \eta};\bm{\theta})
\sum\limits_{t=1}^T \phi_k(\langle s, a \rangle_t)
\label{eq:ce-phi}
\end{align}

We may employ the Baum-Welch algorithm~\cite{Rabiner89:Tutorial} to find the distributions $Pr(X | {\bm \omega}, {\bm \eta};\bm{\theta})$.  However, due to the large amounts of observation nodes per trajectory we replace the forward-backward message passing of Baum-Welch with Markov chain Monte Carlo sampling to improve performance. As shown in Algorithm 1 {\tt HierarchicalBayesInference} present in the Appendix, we first sample all the local nodes $(S, A, Z)$ to produce distributions over all complete trajectories, update the observation model using these distributions, and repeat until convergence.  
To sample nodes we employ Metropolis-within-Gibbs sampling~\cite{tierney1994markov}, a hybrid technique similar to Gibbs sampling, which samples individual nodes one at a time.  {\em However, when the transition function of the underlying MDP is deterministic, we may not efficiently sample one state or action node at a time due to the probability of all transitions but one being zero.}  Then, we modify our algorithm slightly to sample a trajectory's entire set of state-action nodes at once using a method similar to Metropolis-Hastings.

\subsection{Exploiting Indirect Observations}

Although occlusion of the subject is one potential challenge in partially-controlled environments, it may be mitigated by exploiting {\em indirect observations} received from the subject. 

By its nature, the percept that causes an indirect observation does not travel directly from the subject agent to the learner's sensors, rather it reflects or bounces off of some part of the environment, appearing to come from elsewhere than the subject.  These observations could include mirror reflections, natural pinspeck cameras \cite{torralba2012accidental}, shadows~\cite{Baradad_2018_CVPR}, and changes in ambient color. Because they partly depend on the specific demonstration environment, these observations are difficult to manually specify and detect. It is also difficult to simulate them, as they rely on complex calculations such as ray tracing for optical physics, requiring extensive computation time.

As a simple instance of indirect observations countering occlusion, suppose a color blob finder is being used to identify an object being manipulated by a robot.  If a reflective surface exists in the environment, such as the chrome plating of a robot arm, we may receive consistent indirect observations through this surface.  If the subject agent accidentally occludes the object from direct view, it is possible the reflection is still visible and can be used to identify the object.  

Our approach is to allow the observation model, $O$, to be updated from the available demonstration data.  This can be seen in line 23 of Algorithm 1 (see the Appendix), where we gradually update $\alpha^i$ using $\dot{\alpha}$ found using the expected trajectories at iteration $i$ and the previous values. In occluded timesteps, the number of observations directly caused by the subject agent is small, if any. Thus, any indirect observations greatly influence the resulting distribution over the expert's states and actions in these timesteps.

\section{Experiments}

We implement the algorithm and evaluate it on the toy Gridworld as well as in our domain of key interest -- the sorting of vegetables using a robot manipulator. 

\subsection{Formative Evaluation on Gridworld}

We first evaluate our model on a 5$\times$5 Gridworld MDP.  The Gridworld's reward function has four features  corresponding to being in each of the four corners, with one corner randomly chosen to be the goal. We define four observations corresponding to each of the corners. The true observation function has each occurring exponentially more likely as the agent gets closer to the respective corner. For more details about the Gridworld, see the Appendix.

To this model we add some confounding elements, each with its own random observation function.  For our experiment we simulate the subject agent moving through the Gridworld 20 times, and at each timestep produce 40 observations sampled equally from the subject agent's and any confounding element's observation function.  Thus, a the number of confounding elements increases, the proportion of observations attributed to the subject reduces, making this a difficult learning task.

\vspace{-0.15in}
\begin{figure}[!ht]
	\centering
	\includegraphics[width=0.425\textwidth]{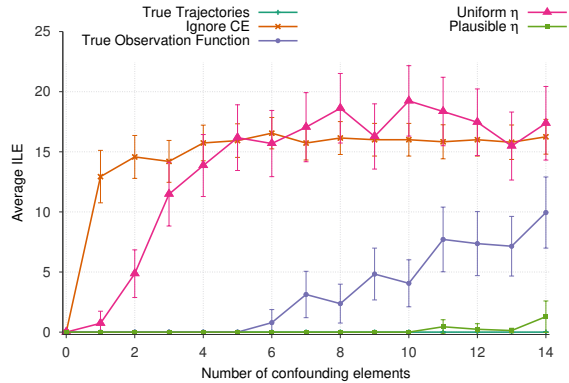}
	\caption{\small Average ILE of various techniques as the \# of confounding elements increases in Gridworld. Each data point is the result of 65+ runs. Error bars are 95\% confidence intervals on the mean.}
	\label{fig:gridworldgraph}
	\vspace{-0.05in}
\end{figure}

\begin{figure*}[!ht]
	\centering
	\begin{minipage}{0.2\textwidth}
	\includegraphics[width=0.475\textwidth]{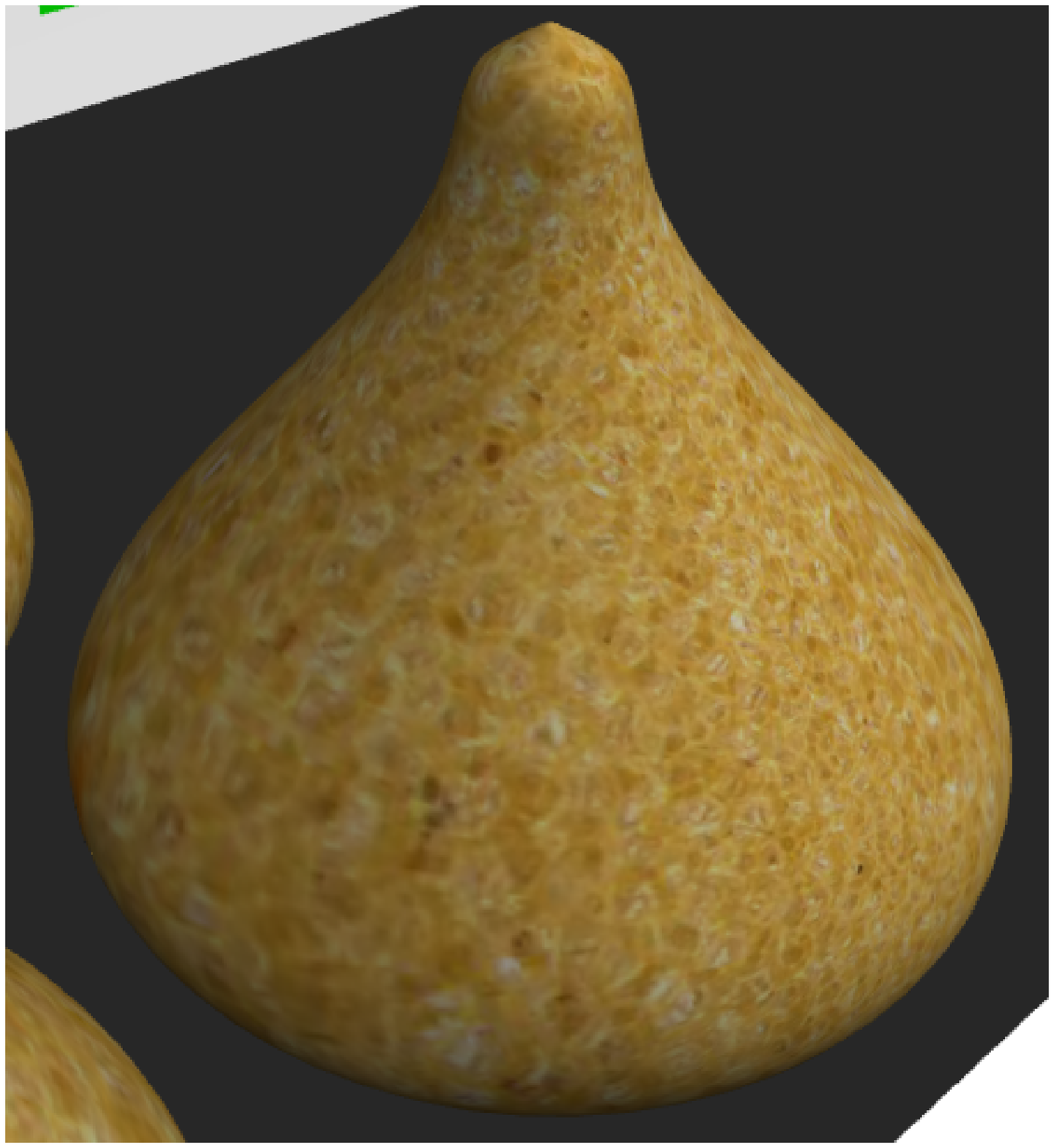}
	\includegraphics[width=0.475\textwidth]{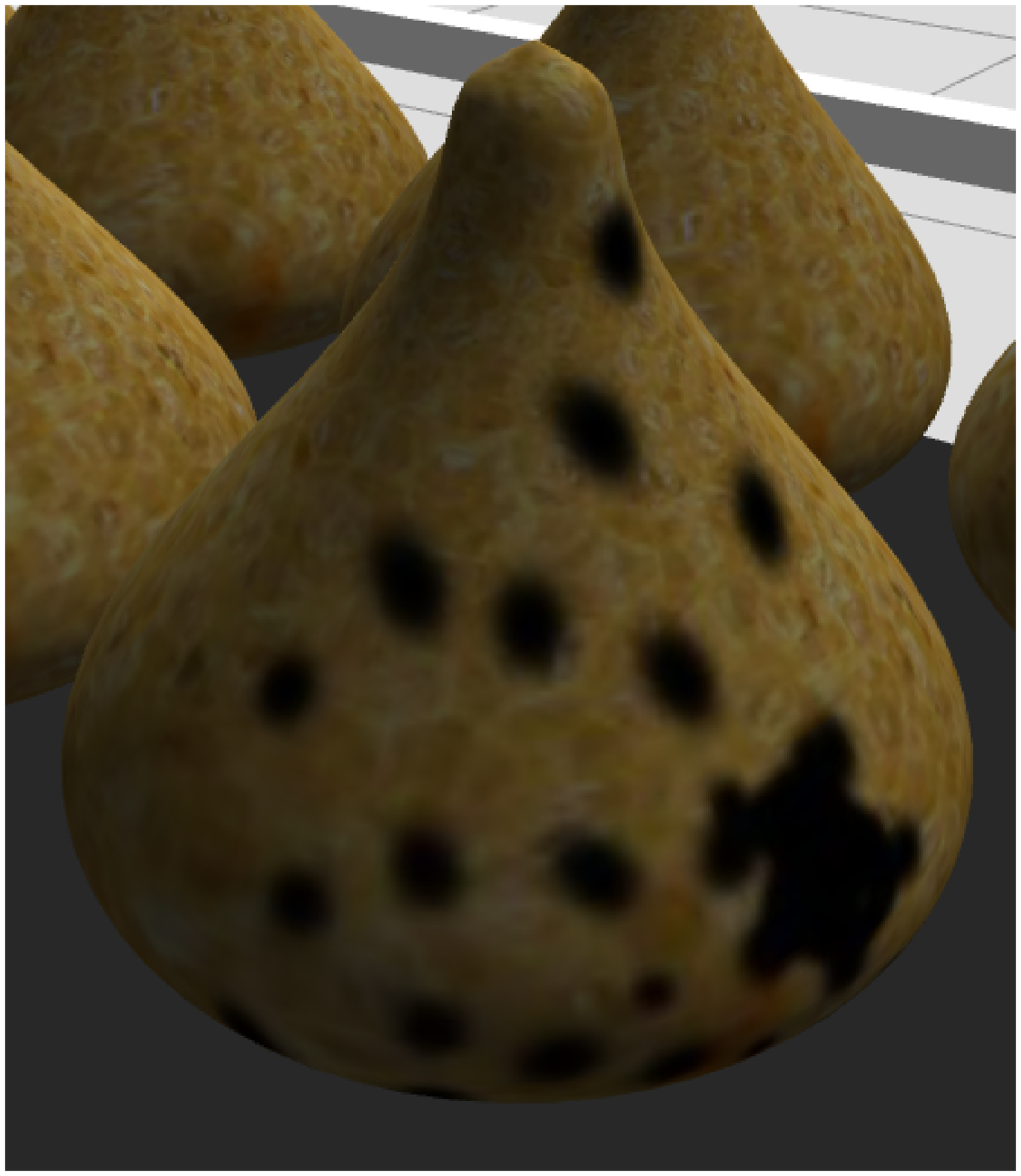}
	\centerline{\small $(a)$}
	\end{minipage}
	\hspace{0.25in}
	\begin{minipage}{0.15\textwidth}
	\includegraphics[width=\textwidth]{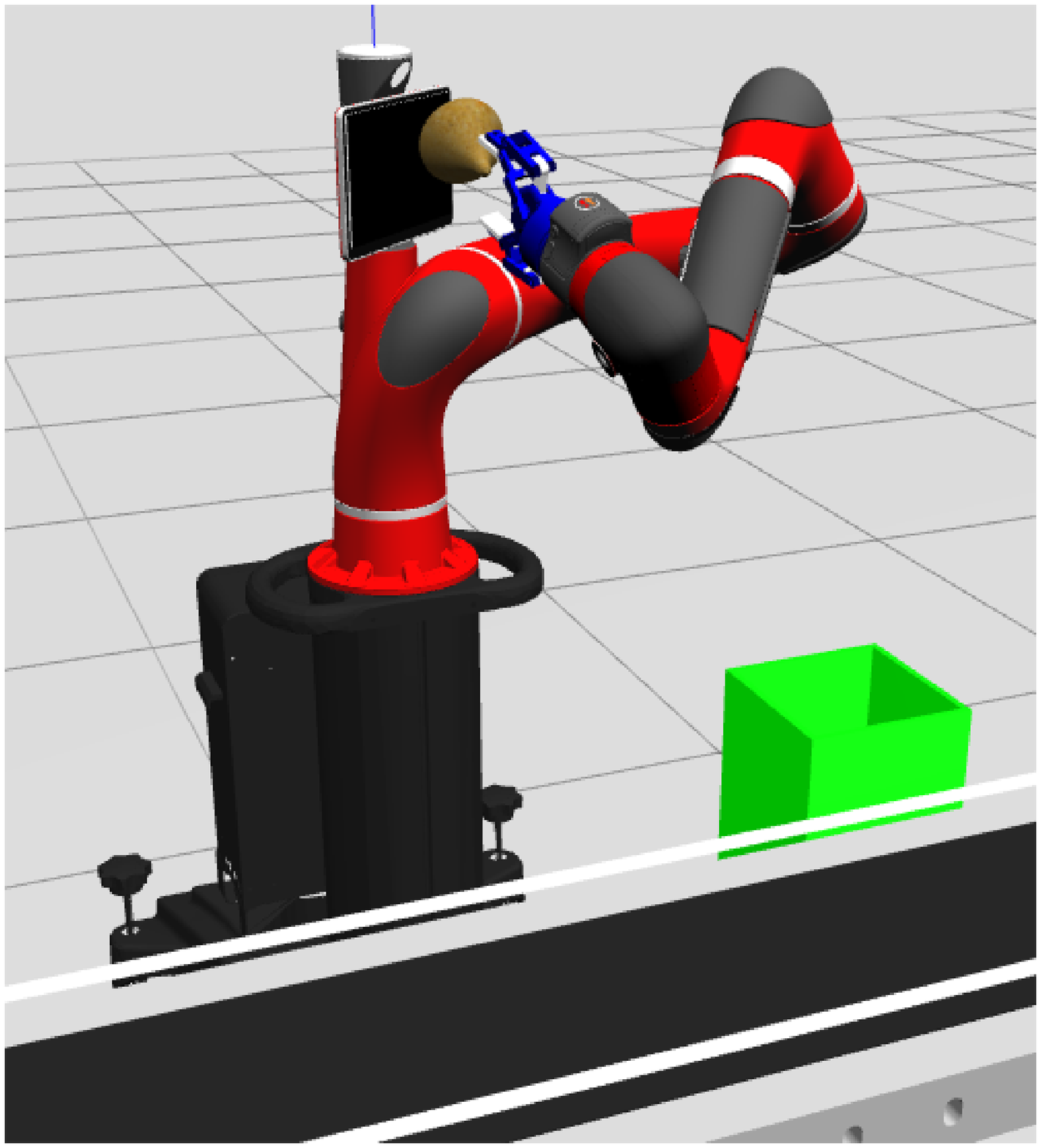}
	\centerline{\small $(b)$}
	\end{minipage}
	\hspace{0.25in}
	\begin{minipage}{0.155\textwidth}
	\includegraphics[width=\textwidth]{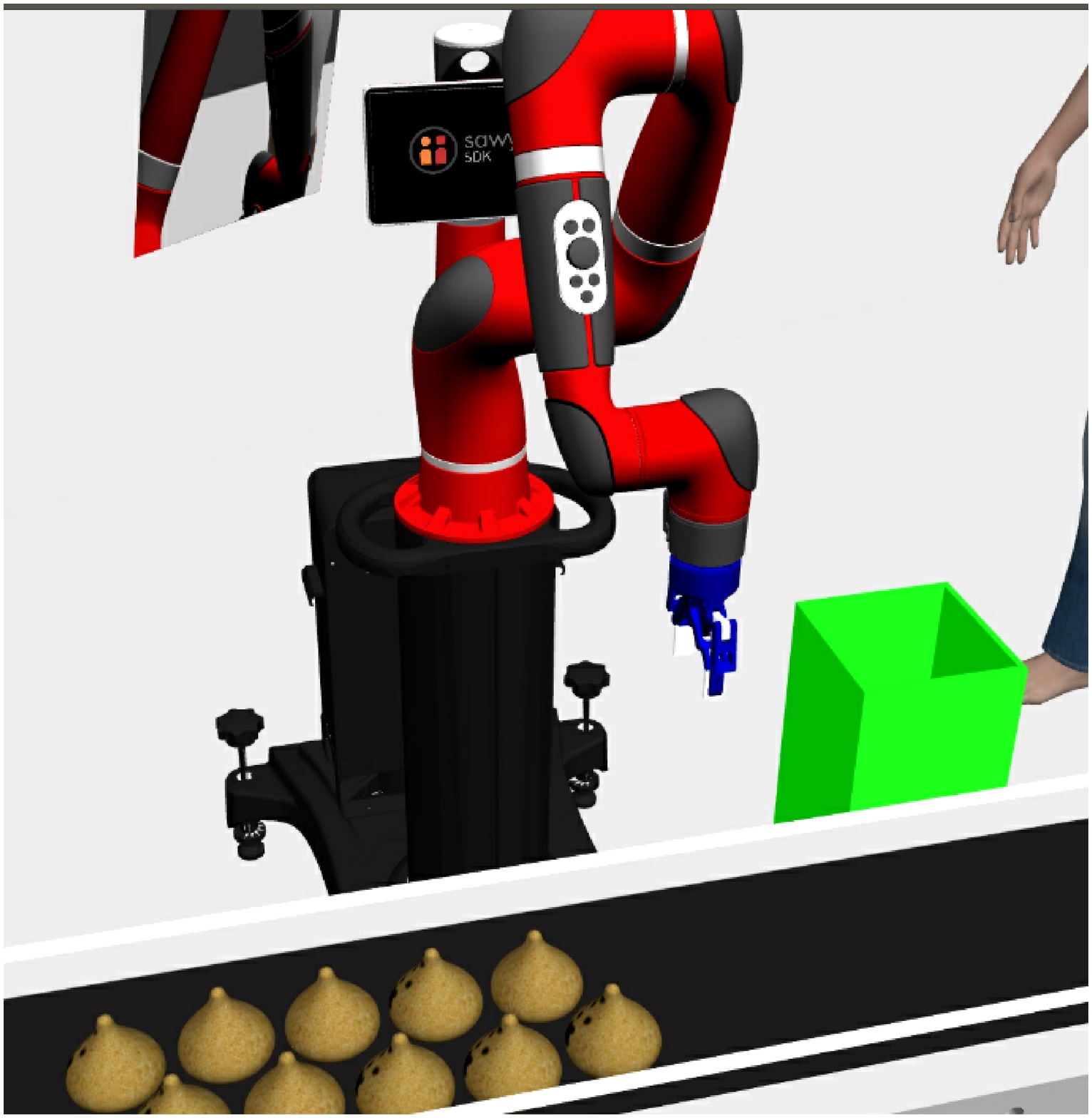}
	\centerline{\small $(c)$}
	\end{minipage}
	\begin{minipage}{0.155\textwidth}
	\includegraphics[width=\textwidth]{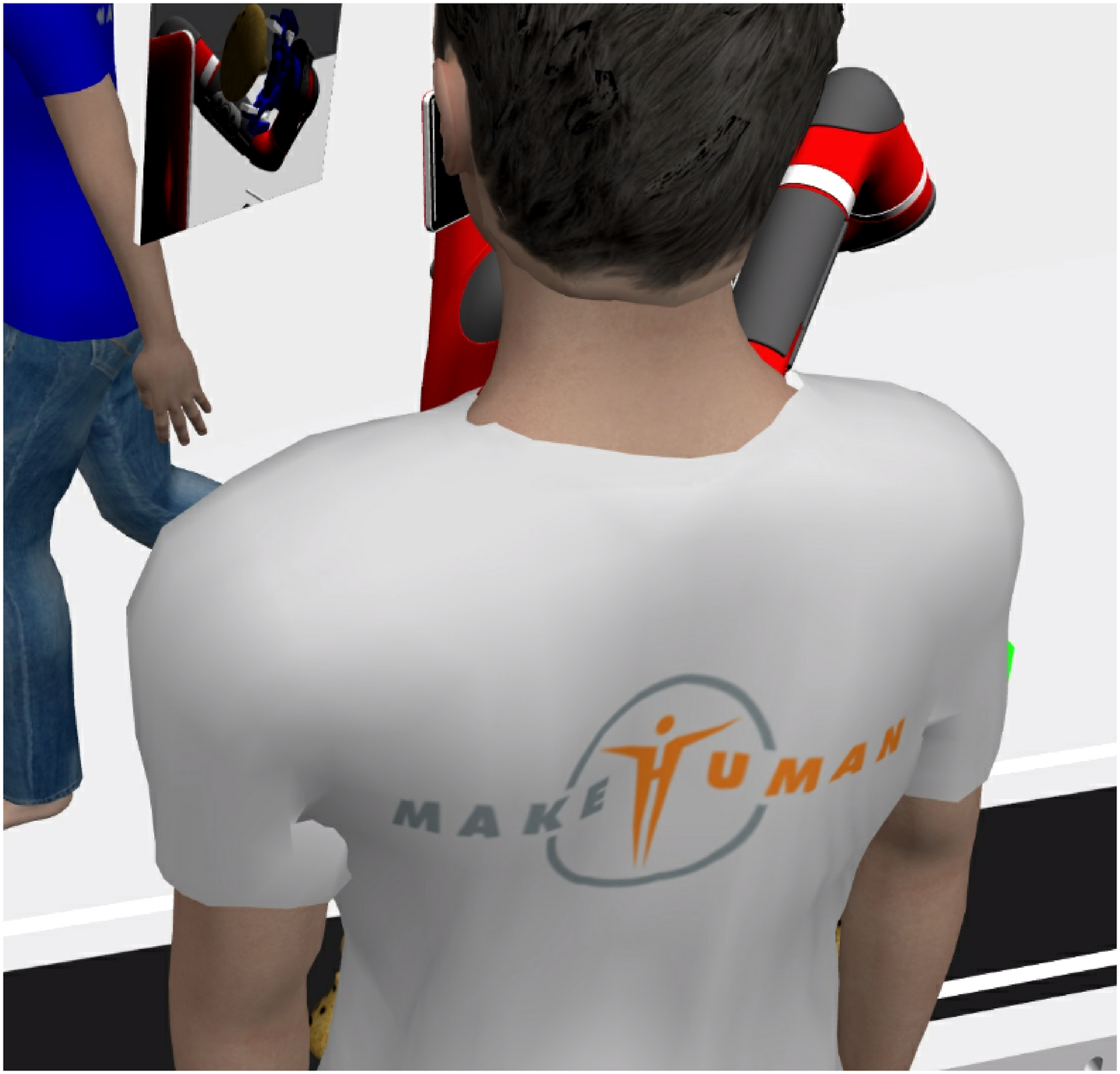}
	\centerline{\small $(d)$}
	\end{minipage}
	\begin{minipage}{0.155\textwidth}
	\includegraphics[width=\textwidth]{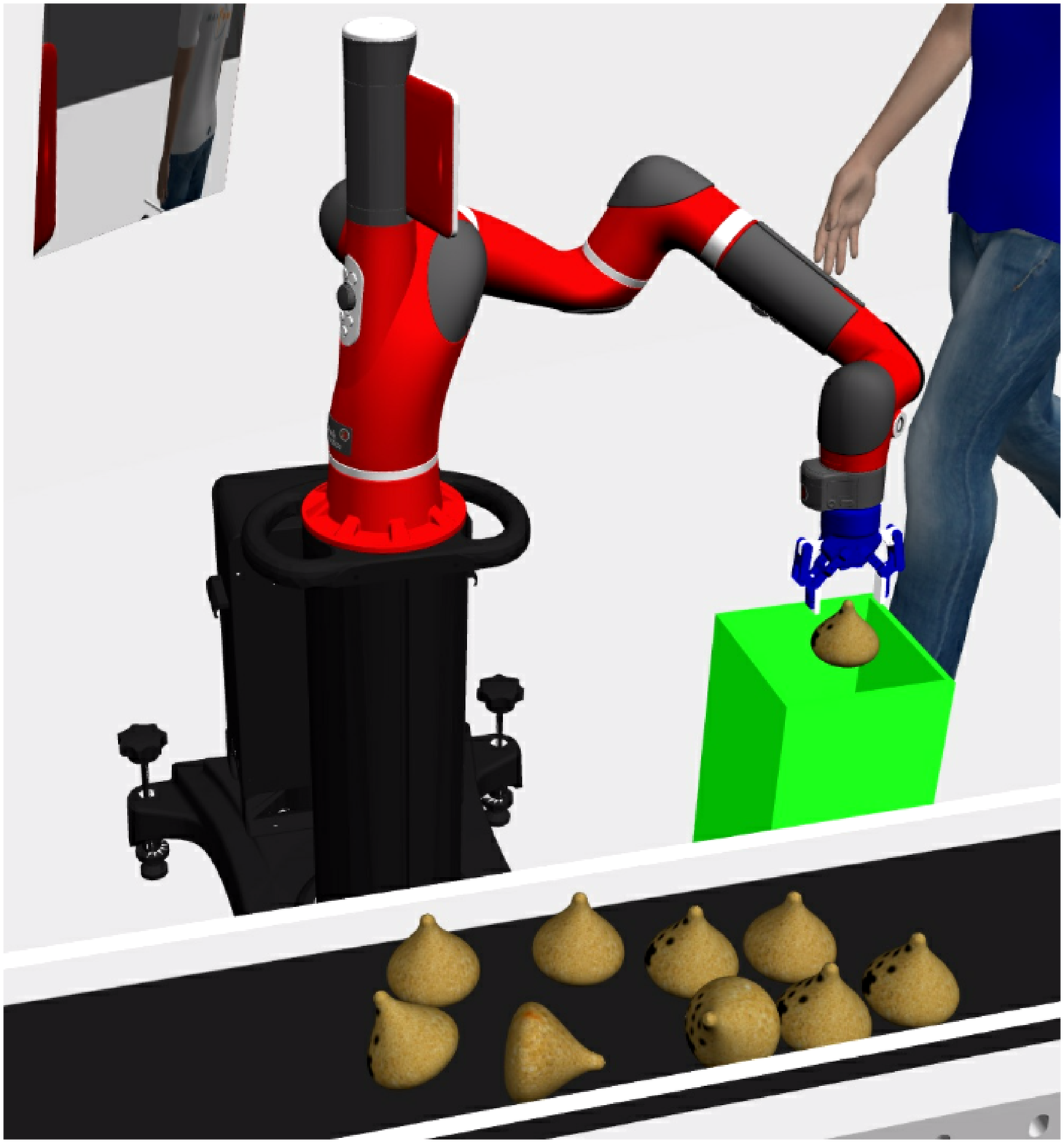}
	\centerline{\small $(e)$}
	\end{minipage}
	\caption{\small ($a$) Blemished onions from two viewpoints. ($b$) The onion sorting robot inspecting a bad onion in a fully controlled environment. ($c$,$d$,$e$) The same robot in a partially controlled environment sorting onions.  Note the blue gripper and an individual's blue shirt, occlusion, presence of other onions, and the mirror, which provides the indirect observations (see the supplemental simulation video for more details).}
	\label{fig:experimentscreenshots}
\vspace{-0.2in}
\end{figure*}

Figure~\ref{fig:gridworldgraph} shows the ILE of several variants of our algorithm.  First,  $\mathbf{True~Trajectories}$ where the subject's true trajectories are input into MaxCausalEntropyIRL~\cite{Ziebart2010}, has 0 error as expected.  Next is a variant where the {\bf True}  $\mathbf{Observation~Function}$ $O$ of the subject and confounding elements are known to the learner and all $\eta$ are uniform. It exhibits excellent performance with up to 6 confounding elements.  In the next two variants, the observation model $O$ is unknown but an informative Dirichlet prior $\alpha$ is used for the subject agent and symmetric (uninformative) prior for the confounding elements.\footnote{Informative observation model priors for confounding elements did not significantly impact performance in our experiments.} $\mathbf{Uniform~\eta}$ uses a uniform $\eta$ distribution for all observations while $\mathbf{Plausible~\eta}$ has a 80\% chance of assigning the true source of an observation 0.6 probability mass, otherwise a random source is given this mass.  Finally, \textbf{Ignore CE} treats all observations as if they came from the subject. It achieves 0 error with no confounding elements but ILE rapidly rises as $|\E|$ grows.  Overall, Fig.~\ref{fig:gridworldgraph} shows that our technique with a plausible $\eta$ performs nearly as well as having the true trajectories with up to 13 confounding elements.

\subsection{Summative Assessment using Robotic Onion Sorting}

\noindent {\bf Domain description~~} We now examine the performance of our technique in a larger simulated vegetable sorting task. The Sawyer robot arm is tasked with inspecting onions moving down a conveyor belt and sorting good onions from blemished ones.  A blemished onion exhibits dark portions which may be visible on the conveyor but will generally require a closer inspection by the robot's cameras.  In this pick-inspect-place scenario we desire installing a new robot that will work alongside the existing one to increase sorting capacity.  Rather than manually program the new robot, however, we use apprenticeship learning to have it learn from watching the existing one's behavior in a production environment, such as a warehouse, where many confounding elements are present that cannot be removed. Examples of such elements include warehouse workers, other robots, and stations sorting vegetables other than onions. To assist the learner, we place a mirror behind the sorting robot to give the learner an additional viewpoint into the inspection. When the subject robot is occluded from the learner's view, {\em the mirror offers indirect observations of the onion being inspected and the subject's actions to the learner.} 

Our MDP for the sorting task is discrete. The states are composed of 3 discrete variables: {\sf Onion Quality} (unknown, good, blemished), {\sf Onion Position} (on\_conveyor, gripped, in\_bin), and {\sf Gripper Position} (conveyor, bin, inspection).

Actions cause certain changes in the state. All actions may be taken in any state but will have no effect (state unchanged) other than those described here.  
\begin{itemize}[leftmargin=*,itemsep=0in,topsep=0in]
	\item {\em Grip onion} changes an onion's position from on\_conveyor to gripped with probability 1;
	\item {\em Release onion} changes an onion's position from gripped to in\_bin or on\_conveyor (dependent upon the gripper's position) with probability 1;
	\item {\em Inspect onion} changes an onion's quality from unknown to either good or blemished by closely showing all sides of the onion to the robot's camera (only applicable in the inspection gripper position);
	\item {\em Move to conveyor}, {\em Move to inspection}, {\em Move to bin} changes the gripper's position to conveyor, inspection, and the bin each with probability 1, respectively.
\end{itemize}
	
Four binary {\bf reward features} are defined below and take the value 0 in all states and actions except in those described where they are valued 1.
\begin{enumerate}[leftmargin=*,itemsep=0in,topsep=0in]
	\item Release good onion in bin;
	\item Release good onion back on the conveyor belt;
	\item Release blemished onion in bin;
	\item Release blemished onion back on the conveyor belt.
\end{enumerate}
The MDP assumes the subject is placing each type of onion either in the bin or back on the conveyor. {\em The learner is unaware of which action is chosen for which onion type.}

Trajectories are short (5-6 timesteps long) and terminate when Sawyer performs a {\em Release onion} action, thereby completing a single sort. Timesteps correspond to 2 secs of wall-clock time and we assume the learner is able to detect the beginning and end of trajectories (or an operator indicates them).

Our learner robot is equipped with a fixed RGB camera that is observing the nearby subject. Determining the subject's complete state is not straightforward: Consider a blemished onion as shown in Fig.~\ref{fig:experimentscreenshots}$(a)$.  It appears identical to a good onion from one angle, but from others has detectable dark spots.  These appear to the learner as a distribution of discrete observations: bright onion and dark onion, with blemished onions producing relatively more dark onion observations.

Additionally, confounding elements are present in the partially-controlled trials. These include moving persons in the environment that cause false positive observations or occlude the subject from view, and other onions not being sorted (see Fig.~\ref{fig:experimentscreenshots}$(c,d,e)$ and the supplemental video). The observations are defined below with more details in the Appendix:
\begin{itemize}[leftmargin=*,itemsep=0in,topsep=0in]
	\item Gripper blobs - A color blob finder is run on RGB images to detect the gripper.  Gripper positions produce distributions of color blobs which are classified into one of 20 discrete observations each corresponding to an equal-sized region of the camera's f.o.v. The $\eta$ for these observations is based on the size of the blob, with more mass given to the subject as the blob size approaches the bounding box size of the gripper in pixels.  The leftover mass is attributed to a person walking behind the expert wearing a shirt of the same color as the gripper.
	
	\item Bright onion, Dark onion - Onions are detected in the RGB images using a color blob finder.  The luminosity of the pixels making up the onion are then classified as either Bright or Dark based on the proportion of dark pixels present.  $\eta$ is set to 80\% for subject if the blob is proximal to a gripper blob, with the remainder going to other onions and a foreground person. If no gripper blob is seen nearby, we assign 60\% mass to the other onions for small blobs or a foreground person for large ones, with the remainder divided between the subject (10\%) and the other confounding element (25\%).
\end{itemize}
Notice that these observations do not unambiguously specify the state and action of the subject robot.  Additionally, the $\eta$ values chosen represent our uncertain estimate of the source of the detected blobs, illustrating the difficulty of filtering out confounding element observations.

\vspace{0.05in}
\noindent {\bf Experiment procedure~~} We simulate the sorting robot with all confounding elements removed and record $(i)$ the true state and action in each timestep and $(ii)$ all observations received.  We use these to find the $\alpha$ hyperparameters, as well as provide a ground truth trajectory for upper bound purposes.

Next, we repeat the same sorting procedure but with a mirror installed and confounding elements added, recording all observations received.  We produce the following data sets:
\begin{itemize}[leftmargin=*,itemsep=0in,topsep=0in]
    \item Maximum likelihood (ML) trajectories - given the observation model and observations received we choose the most likely expert state and action each timestep.  This approximates common practice and serves as a baseline.  
    \item ML observations - given $\eta$ values as defined above, choose the observations that are most likely caused by the subject and use these observations only in the ``controlled environment''.  This demonstrates the accuracy of $\eta$ distributions.
    \item Partially-controlled model - we define the set $\E$ manually by reviewing the recorded procedure and observation stream and listing all objects that cause observations other than the subject.  We use uninformative observation hyperparameters $\alpha$ for all confounding elements. We then use the observations in our partially controlled model.  $\eta$ is either  $\mathbf{Plausible}$ (set to the values as described previously) or is $\mathbf{Uniform}$.
\end{itemize}

We utilize MaxCausalEntIRL to find a reward function for data sets with known states and actions.  For all others, we use HiddenDataEM and report metrics after Algorithm 1 has converged. We show the ILE produced by using the learned reward function to complete the given MDP versus using the true reward function for each data source. There were three confounding elements, which are evident in Fig.~\ref{fig:experimentscreenshots}$(c,d)$.

\vspace{-0.15in}
\begin{figure}[!ht]
	\centering
	\includegraphics[width=0.45\textwidth]{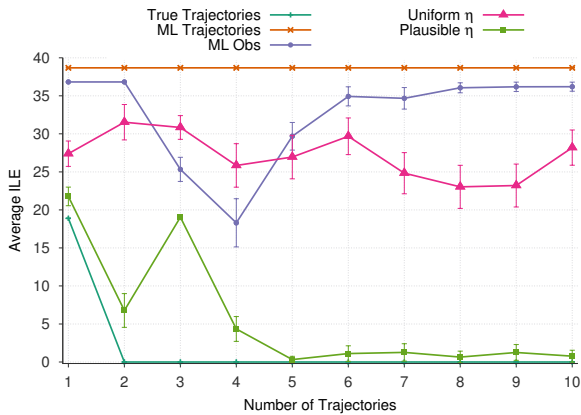}
	\caption{\small Average ILE in the onion sorting experiment as the number of trajectories increases. Each data point is from 100+ runs. Error bars are 95\% confidence intervals on the mean.}
	\label{fig:onionsortgraph}
	\vspace{-0.1in}
\end{figure}

As can be seen in Fig.~\ref{fig:onionsortgraph}, two trajectories are needed for \textbf{True Trajectories} to achieve zero error due to the need to see at least one good and one blemished onion being sorted.  \textbf{Plausible $\eta$} achieves very low error with 4 or more trajectories, while all other techniques fail to reduce error as trajectories increase. \textbf{ML Trajectories} demonstrated high sensitivity to the trajectories due to the variance in observations each trajectory produces.  As another verification we apply our techniques to observations produced in the fully controlled scenario and   all methods can achieve 0 error with just two trajectories.

%-----------------------------------------------------------------------------------------------------------
\section{Related Work}
\label{sec:related}
%-----------------------------------------------------------------------------------------------------------

Bogert and Doshi~\cite{Bogert14:Multi,Bogert18:Multi} introduces the challenges of occlusion that manifest in real-world applications of IRL to robotics. This extension of the original problem of IRL -- which assumes full and perfect observability of the trajectories -- is formally defined in a recent survey of the methods and progress in IRL~\cite{Arora2018survey}. An initial method~\cite{Bogert14:Multi} extended MaxEnt to settings involving occlusion of portions of the trajectory by limiting  the constraints to the observed  portions only. 
Subsequently, the EM based approach reviewed in Section~\ref{sec:background} was introduced, which forms an  expectation over the missing data to allow the  use of the Lagrangian  gradient.  However, this  method  suffers from  a computationally intensive expectation step when the occluded portions of the trajectories are long and contiguous. This limitation is addressed by a follow-up paper~\cite{Bogert:2017:Scaling}, which shows how blocked Gibbs sampling can be utilized to speed up the forward-backward message passing in the expectation step to allow scaling under occlusion to multiple experts. However, none of these generalizations to occlusion integrate indirect observations in the model as we do in this paper, which represents a different approach to manage occlusion.

Kitani et. al.~\cite{Kitani2012} introduce IRL with a single observation per timestep of the subject agent. However, the observation is modeled as caused by the subject's state only and no mechanism for exploiting the learned policy to improve the state distribution is developed. Choi and Kim~\cite{Choi15:Hierarchical} generalized Bayesian IRL~\cite{Ramachandran07:Bayesian} to a hierarchical model by introducing a prior over the temperature parameter that determines the randomness of the softmax distribution used in Bayesian IRL, and a hyperprior over the reward function prior distribution. Maximum a-posteriori inference for Bayesian IRL~\cite{Choi2013bayesian} is then extended to the hierarchical version with a demonstration on the Gridworld and taxi problems~\cite{Ziebart08:Maximum}. A different hierarchical Bayesian model was also introduced by Dimitrakakis and Rothkopf~\cite{Dimitrakakis12:Bayesian} to generalize Bayesian IRL to learning multiple reward functions. The model involves a hyperprior on a joint reward-policy distribution for each of the demonstrated tasks. The posterior distributions are obtained using a Metropolis-Hastings based sampling technique, which is demonstrated on simple random MDPs. While these two methods involve the use of hierarchical models, these are relatively shallow compared to the final model we utilize here and are not immediately generalizable to address the challenges of occlusion.

\section{Concluding Remarks}

We relax the strong assumption in IRL that the expert's true trajectories are known to the learner. As demonstrated in our experiments involving a robotic domain, we successfully generalize a previous method to realistic environments where confounding elements introduce noisy observations.  Our method is shown to be resilient to error in the source distribution $\eta$. Future work will explore our technique's sensitivity to $\eta$, allow this distribution to be learned from data, and extend the technique to uncontrolled scenarios where the set of confounding elements is unknown.

%------------------------------------------------------------------------------------
\bibliographystyle{IEEEtran}
\bibliography{bdIROS21}

%{\setlength{\parindent}{0pt}
\section*{Technical Appendix}

\subsection{Algorithms}

\begin{algorithm}[!ht]
\SetAlgoLined
\SetKwInOut{Input}{Input}
\DontPrintSemicolon
\begin{small}

\Input{$\mathbb{O}$, $\alpha^0, \bm{\theta}$}
\KwResult{$Pr(X | \bm {\omega}, \bm {\eta)}$}
$Pr(a|s) \leftarrow SoftMaxPolicy(\bm{\theta}) ~~ \forall s,a$\;
$O_{s,a} \leftarrow Dirichlet\_mean(\alpha_{s,a}^0) ~~\forall~ s,a$\;
$O_{Z} \leftarrow Dirichlet\_mean(\alpha_Z^0) ~~\forall~ Z_{/1}$\;
$i \leftarrow 0$\;
\While{Not all $O$ converged}{

    $i \leftarrow i + 1$\;
    \For {${\bm \omega}, {\bm \eta} \in \mathbb{O}$} {
        Initialize \{ $Z_{t,n}^0 : t = 1, ... T, n = 1 ... N^t$\}, \{ $s_t^0 : t = 1, ... T$\}, \{ $a_t^0 : t = 1, ... T$\} \\
        \For{$j\leftarrow 1$ \KwTo $\sigma$} {
            \For {$t \leftarrow 1$ \KwTo $T$} {
                \For {$n \leftarrow 1$ \KwTo $N^t$} {
                Sample $Z_{t,n}^{(j)} \sim Pr(Z_{t,n} | Z_{t,n}^{(j-1)},$ $\eta_{t,n}, \omega_{t,n}, O, s_t^{(j-1)}, a_t^{(j-1)})$\;
                }
                Sample $s_t^{(j)} \sim  Pr(s_t | s_t^{(j-1)}, a_t^{(j-1)}, s_{t+1}^{(j-1)}, s_{t-1}^{(j-1)},$ $a_{t-1}^{(j-1)}, \Omega_t,$\\ ~~~~~~~$Z_t^{(j)}, O) $\;
                Sample $a_t^{(j)} \sim Pr(a_t | s_t^{(j)}, s_{t+1}^{(j)}, \Omega_t, Z_t^{(j)}, O) $\;
                
            }
        }
        Estimate $Pr(Z_{t,n} | {\bm \omega}, {\bm \eta})$ from $Z_{t,n}^{(1 ... \sigma)} ~~\forall~ Z, t, N^t$\;
        Estimate $Pr(X | {\bm \omega}, {\bm \eta})$ from $s^{(1 ... \sigma)}, a^{(1 ... \sigma)} ~~\forall~ X$\;
    }
    % need to redo these calculations in terms of $Pr(X | \Omega, \eta) and Pr(Z_t,n | \Omega, eta)
    $\dot{\alpha}_{s,a} \leftarrow \sum\limits_{\langle {\bm \omega}, {\bm \eta} \rangle \in \mathbb{O}} \sum\limits_{t \in {\bm \omega}} \sum\limits_{n=1}^{|{\bm \omega}_{t}|} Pr(Z_{t,n} = 1 |  {\bm \omega}, {\bm \eta}) \sum\limits_{X : s_t = s, a_t = a} Pr(X | {\bm \omega}, {\bm \eta}) ~~\forall ~s,a$\;
    $\dot{\alpha}_{Z} \leftarrow\sum\limits_{\langle {\bm \omega}, {\bm \eta} \rangle \in \mathbb{O}} \sum\limits_{t \in {\bm \omega}} \sum\limits_{n=1}^{|{\bm \omega}_{t}|} Pr(Z_{t,n} = Z | {\bm \omega}, {\bm \eta}) ~~\forall ~Z_{/ 1}$\;

    $\alpha^i \leftarrow c~\dot{\alpha} + (1-c)\alpha^{i-1} ~~\forall~\alpha$\;

    $O_{s,a} \leftarrow Dirichlet\_mean(\alpha_{s,a}^i) ~~\forall~ s,a$\;
    $O_{Z} \leftarrow Dirichlet\_mean(\alpha_Z^i) ~~\forall~ Z_{/ 1}$\;

}
\end{small}
\caption{\sc HierarchicalBayesInference}
\label{alg:main}
\end{algorithm}

where $c$ is a small constant, $\sigma$ is the number of desired samples, $Dirichlet\_mean()$ computes the mean distribution for a Dirichlet with the given hyperparameters, and 
\begin{flalign*} 
&SoftMaxPolicy(\bm {\theta}) \triangleq Pr(a|s) ~=~ e^{ (Q^{soft}(s, a) - V^{soft}(s))}\\ &~\forall~ s,a\\
&Q^{soft}(s_t, a_t) ~=~ E_{s_{t+1}} \left [ V^{soft}(s_{t+1}) | s_t, a_t \right ]\\
&V^{soft}(s_t) ~=~ \softmax\limits_{a_t} \left [ Q^{soft}(s_t, a_t) + \bm{\theta}^T \phi(s_t, a_t) \right ]
\end{flalign*}
Initialization procedure, not shown, involves finding an initial non-zero-probability trajectory and sampling repeatedly to reduce the impact of this initial state on the final result.

\begin{algorithm}[!ht]
\SetAlgoLined
\SetKwInOut{Input}{Input}
\DontPrintSemicolon

Given $y^{\tau - 1}, \zeta ~=~ $all other current node values

Simulate $\tilde{y} \sim U(Y)$\;

Take $y^\tau = \begin{cases}
    \tilde{y}& \text{with probability}~~ p\\
    y^{\tau-1}& \text{with probability} ~~1 - p
\end{cases}$\\

Where: $p ~=~ \min \left ( 1, \frac{\displaystyle Pr(\tilde{y} ~|~\zeta)} {\displaystyle Pr(y^{\tau - 1} ~|~\zeta)} \right )$

\caption{\sc Sample}
\label{alg:sample}
\end{algorithm}

Where the probabilities of individual nodes ($y$) are defined as:

\begin{flalign*} 
&Pr(Z_{t,n} | \eta_{t,n}, \omega_{t,n}, O, s_t, a_t)=\\ 
&\begin{cases}
    \eta_{t,n}(Z_{t,n}) O_{s_t, a_t } (\omega_{t, n}),& \text{if }  Z_{t,n} = 1\\
    \eta_{t,n}(Z_{t,n}) O_{Z_{t,n} } (\omega_{t, n}),              & \text{otherwise}
\end{cases}\\
& Pr(s_t | a_t, s_{t+1}, s_{t-1}, a_{t-1}, \Omega_t, Z_t, O) ~=~ Pr( a_t | s_t)\\
&\times Pr(s_t | s_{t-1}, a_{t-1})~Pr(s_{t+1} | s_t, a_t)\prod\limits_{n : Z_{t,n} = 1} O_{ s_t, a_t } (\omega_{t, n})\\
& Pr(a_t | s_t, s_{t+1}, \Omega_t, Z_t, O) ~=~ Pr( a_t | s_t) Pr(s_{t+1} | s_t, a_t)\\
& \times \prod\limits_{n : Z_{t,n} = 1} O_{s_t, a_t } (\omega_{t, n})
\end{flalign*}

\textbf{Implementation note:} When a large number of observations are present in a timestep the multiplication of the $Z$ nodes in $Pr(s_t)$ and $Pr(a_t)$ will not be numerically stable. 
In this situation, we modify the computation of the ratio $ Pr(\tilde{y}) / Pr(y^{\tau - 1})$  to use a sum of log probabilities, ie. for $s_t$: $exp ( \dots + \left ( log~ O_{ \tilde{s}_t, a_t } (\omega_{t, n}) - log~ O_{ s_t^{\tau - 1}, a_t }(\omega_{t, n}) \right ) + \dots )  ~~\forall~ n : Z_{t,n} = 1$.
%}

\end{document}